\documentclass[10pt,twocolumn,letterpaper]{article}

\usepackage{cvpr}
\usepackage{times}
\usepackage{epsfig}
\usepackage{graphicx}
\usepackage{amsmath}
\usepackage{amssymb}

\usepackage{helvet} % DO NOT CHANGE THIS
\usepackage{courier}  % DO NOT CHANGE THIS
\usepackage[hyphens]{url}  % DO NOT CHANGE THIS
\usepackage{graphicx} % DO NOT CHANGE THIS
\usepackage[utf8]{inputenc} % allow utf-8 input
\usepackage[T1]{fontenc}    % use 8-bit T1 fonts
\usepackage{url}            % simple URL typesetting
\usepackage{booktabs}       % professional-quality tables
\usepackage{amsfonts}       % blackboard math symbols
\usepackage{nicefrac}       % compact symbols for 1/2, etc.
\usepackage{float}
\usepackage{graphicx}
\usepackage{algorithm}
\usepackage{algorithmic}
\usepackage{xcolor}
\usepackage{subfigure}
\usepackage{multirow}
\usepackage{wrapfig}
\usepackage{amsthm}

\newcommand{\aka}{a.k.a. }

\graphicspath{{../}}

\theoremstyle{definition}
\newtheorem{definition}{Definition}

\usepackage[breaklinks=true,bookmarks=false]{hyperref}

\cvprfinalcopy

\def\cvprPaperID{2057} % *** Enter the CVPR Paper ID here

\ifcvprfinal\fi
\begin{document}
	
	%%%%%%%%% TITLE
	\title{CARS: Continuous Evolution for Efficient Neural Architecture Search}
	
	\author{Zhaohui Yang$^{1}$, Yunhe Wang$^{2}$, Xinghao Chen$^{2}$, Boxin Shi$^{3,4}$,\\
	Chao Xu$^{1}$, Chunjing Xu$^{2}$, Qi Tian$^{2}$\thanks{Corresponding author.}, Chang Xu$^{5}$ \\
		\normalsize$^1$ Key Lab of Machine Perception (MOE), Dept. of Machine Intelligence, Peking University.
\\
		\normalsize$^2$ Huawei Noah's Ark Lab. \normalsize$^3$ NELVT, Dept. of CS, Peking University. \normalsize$^4$ Peng Cheng Laboratory. \\
		\normalsize$^5$ School of Computer Science, Faculty of Engineering, University of Sydney.\\
		\small\texttt{\{zhaohuiyang,shiboxin\}@pku.edu.cn; xuchao@cis.pku.edu.cn}\\
		\small\texttt{\{yunhe.wang,xinghao.chen,tian.qi1,xuchunjing\}@huawei.com; c.xu@sydney.edu.au}
	}
	
	\maketitle
	%\thispagestyle{empty}
	
	%%%%%%%%% ABSTRACT
	\begin{abstract}
		Searching techniques in most of existing neural architecture search (NAS) algorithms are mainly dominated by differentiable methods for the efficiency reason. In contrast, we develop an efficient continuous evolutionary approach for searching neural networks. Architectures in the population that share parameters within one SuperNet in the latest generation will be tuned over the training dataset with a few epochs. The searching in the next evolution generation will directly inherit both the SuperNet and the population, which accelerates the optimal network generation. The non-dominated sorting strategy is further applied to preserve only results on the Pareto front for accurately updating the SuperNet. Several neural networks with different model sizes and performances will be produced after the continuous search with only 0.4 GPU days. As a result, our framework provides a series of networks with the number of parameters ranging from 3.7\emph{M} to 5.1\emph{M} under mobile settings. These networks surpass those produced by the state-of-the-art methods on the benchmark ImageNet dataset.
	\end{abstract}
	
	\begin{figure}[t]
		\centering
		\begin{tabular}{cc}
			\includegraphics[width=0.9\linewidth]{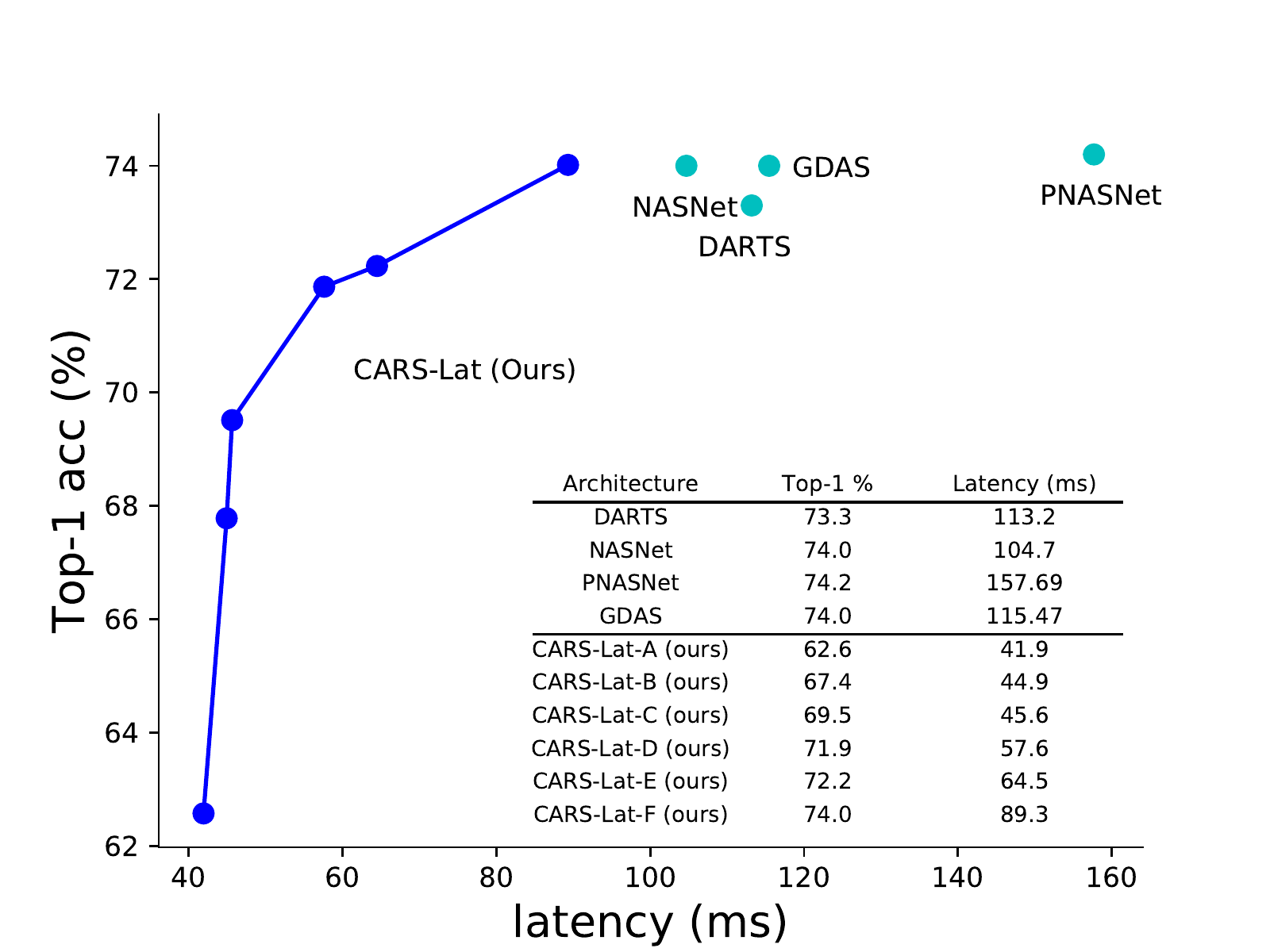}&
		\end{tabular}
		\caption{CARS-Lat models are searched on CIFAR-10 dataset. The search phase considers validation performance and device-aware variable, \ie, mobile device latency on HUAWEI P30 Pro. The Top-1 accuracies are the performances on the ILSVRC2012 dataset.}
		\label{fig_cars_darts_latency}
		\vspace{-1em}
	\end{figure}
	
	\section{Introduction}
	Convolutional neural networks have made great progress in a large range of computer vision tasks, such as recognition~\cite{resnet,ghostnet,aap}, detection~\cite{frcnn}, and segmentation~\cite{maskrcnn}. Over-parameterized deep neural network can produce impressive performance but will consume huge computational resources at the same time. The efficient block design~\cite{slimmable, universallyslimmable, autoslim}, tensor decomposition~\cite{lin2019holistic, fsnet, legonet, wang2018learning, wang2016cnnpack}, pruning~\cite{lin2018accelerating, lin2019towards, targetdropout}, distillation~\cite{datafree, pucompression} and quantization~\cite{bnn} are popular techniques to make networks efficient. Designing novel network architectures heavily relies on human experts' knowledge and experience, and may take many trials before achieving meaningful results~\cite{resnet}. To accelerate this process and make it automated, network architecture search~(NAS)~\cite{mnasnet, autogan, autoreid} has been proposed, and the learned architectures have exceeded those human-designed architectures on a variety of tasks. However, these searching methods usually require lots of computational resources to search for architectures of acceptable performance.
	
	Techniques for searching neural architectures are mainly clustered into three groups, \ie, Evolution Algorithm~(EA) based, Reinforcement Learning~(RL) based, and gradient-based methods. EA-based works, ~\cite{largescaleevolution,amoebanet,geneticcnn,residualdenseblockforsr,sab, coevolution} initialize a set of models and evolve for better architectures, which is time-consuming, \eg, Real \etal takes 3150 GPU days for searching~\cite{largescaleevolution}. RL-based works, \cite{rlnas,blocknas} use the controller to predict a sequence of operations, and train different architectures to gain the rewards. Given an architecture, these methods have to train it for a large number of epochs, and then evaluate its performance to guide for evolution or optimize the controller, which makes the searching stage less efficient. Gradient-based methods (\eg, DARTS~\cite{darts}) first train a SuperNet and introduce the attention mechanism on the connections while searching, then remove weak connections after searching. This phase is conducted by gradient descent optimization and is quite efficient. However, the searched architectures suffer from lack of variety. 
		
	Although some experiments in~\cite{amoebanet} show that the evolutionary algorithm discovers better neural architectures than RL-based approaches, the search cost of EA is much expensive due to the evaluation procedure of each individual, \ie, a neural network in the evolutionary algorithm is independently evaluated. Moreover, there could be some architectures with extremely worse performance in the search space. If we directly follow the weight sharing approach proposed by ENAS~\cite{enas}, the SuperNet has to be trained to compensate for those worse search space. It is necessary to reform existing evolutionary algorithms for efficient yet accurate neural architecture search.
	
	In this paper, we propose an efficient EA-based neural architecture search framework. A continuous evolution strategy is developed to maximully utilize the knowledge we have learned in the last evolution generation. Specifically, a SuperNet is first initialized with considerable cells and blocks. Individuals in the evolutionary algorithm representing architectures derived in the SuperNet will be generated through several benchmark operations (\ie, crossover and mutation). Non-dominated sort strategy is adopted to select several excellent architectures with different model sizes and accuracies, and corresponding cells in the SuperNet will be updated for subsequent optimization. The evolution procedure in the next generation is continuously executing based on the updated SuperNet and the multi-objective solution set obtained by the non-dominated sorting. In addition, we propose to exploit a protection mechanism to avoid the small model trap problem. The proposed continuous evolution architecture search (CARS) can provide a series of models on the Pareto front with high efficiency. The superiority of our method is verified on benchmark datasets over the state-of-the-art methods.

	\section{Related Works}
	
	\subsection{Network Architecture Search}
	
	Gradient-based Network Architecture Search (NAS) methods contain two steps: network parameter optimization and architecture optimization. The network parameter optimization step optimizes the parameters in the standard layers (\ie, convolution, batch normalization, fully connected layer). The architecture optimization step learns the pattern of accurate network architectures.
	
	The parameter optimization step can be divided into two categories, independent optimization and sharing optimization. Independent optimization learns each network separately, \ie, AmoebaNet~\cite{amoebanet} takes thousands of GPU days to evaluate thousands of models. To accelerate training, ~\cite{lemonade,simpleandefficientnas} initialize parameters by network morphism. One-shot methods~\cite{understandingoneshotnas,SPOSNAS} step further by sharing all the parameters for different architectures within one SuperNet. Rather than training thousands of different architectures, only one SuperNet is required to be optimized.
	
	The architecture optimization step includes RL-based, EA-based, and gradient-based approaches. RL-based methods~\cite{rlnas,nasnet,enas} use the recurrent network as the network architecture controller, and the performances of the generated architectures are utilized as the rewards for training the controller. The controller converges during training and finally outputs architectures with superior performance. EA-based approaches~\cite{geneticcnn,amoebanet} search architectures with the help of evolutionary algorithms. The validation accuracy of each individual is utilized as the fitness to evolve the next generation. Gradient-based approaches~\cite{darts,snas,fbnet} view the network architecture as a set of learnable parameters and optimize the parameters by the standard back-propagation algorithm.
	
	\subsection{Multi-objective Network Architecture Search}
	
	Considering multiple complementary objectives, \ie, accuracy, the number of parameters, float operations (FLOPs), energy, and latency, there is no single architecture that surpasses all the others on all the objectives. Therefore, architectures within the Pareto front are desired. Many different works have been proposed to deal with multi-objective network architecture search. NEMO~\cite{nemo} and MNasNet~\cite{mnasnet} target at speed and accuracy. DPPNet and LEMONADE~\cite{dppnet,lemonade} consider device-related and device-agnostic objectives. MONAS~\cite{monas} targets at accuracy and energy. NSGANet~\cite{nsganet} considers FLOPs and accuracy. 
	
	These methods are less efficient for models are optimized separately. In contrast, our architecture optimization and parameter optimization steps are conducted alternatively. Besides, the parameters for different architectures are shared, which makes the search stage much more efficient.
	
	\section{Approach}
	
	In this section, we develop a novel continuous evolutionary approach for searching neural architectures, namely CARS. The CARS search stage includes two procedures, \ie, parameter optimization and architecture optimization.
	
	We use the Genetic Algorithm (GA) for architecture evolution because GA maintains a set of well-performed architectures that cover a vast space. We maintain a set of architectures (\aka connections) ${C} = \{ {C}_1, \dots, {C}_P \}$, where $P$ is the population size. The architectures in the population are gradually updated according to the proposed pNSGA-III method during the architecture optimization step. To make the search stage efficient, we maintain a SuperNet~$\mathcal{N}$, which shares parameters ${W}$ for different architectures. The parameter sharing strategy dramatically reduces the computational complexity of separately training these different architectures.
		
	\subsection{SuperNet of CARS}
	\label{sec_efficient_training}
	
	Different networks are sampled from the SuperNet $\mathcal{N}$, and each network $\mathcal{N}_{i}$ can be represented by a set of full precision parameters ${W}_{i}$ and a set of binary connection parameters (\ie, $\{0,1\}$) ${C}_i$. The 0-element in connection $C_i$ means the network does not contain this connection to transform data flow, and the 1-element connection means the network uses this connection. From this point of view, each network $\mathcal{N}_{i}$ could be represented as $({W}_{i}, {C}_{i})$ pair.
	
	Full precision parameters ${W}$ are shared by a set of networks. If these network architectures are fixed, the parameters could be optimized through back-propagation. The optimal ${W}$ fits for all the networks $\mathcal{N}_{i}$ to achieve higher recognition performance. After the parameters are converged, we could alternately optimize the binary connections ${C}$ by the GA algorithm. These two steps form the main optimization of our proposed method. We will introduce these two optimization steps in the following.
	
	\subsection{Parameter Optimization} 
	
	The parameters ${W}$ are the collection of all the parameters in the network. The parameters ${W}_{i}$ of the $i$-th individual are ${W}_{i} = {W} \odot {C}_{i},~~i\in \{1, \dots, P\}$, where the $\odot$ is the mask operation that keeps the parameters of the complete graph only for the positions corresponding to 1-elements in the connection ${C}_{i}$. Denote ${X}$ as the input data, the prediction of this network is $\mathcal{N}_{i}({X})$, where $\mathcal{N}_{i}$ is the $i$-th architecture. The prediction loss can be expressed as ${L}_i = \mathcal{H}(\mathcal{N}_{i}({X}), Y)$, where $\mathcal{H}$ is the criterion and Y is the target. The gradient of ${W}_{i}$ can be calculated as 
	\vspace{-1em}
	\begin{equation}
	\label{eq_de_wi}
	\begin{split}
	{dW}_{i} &= \frac{\partial{L}_{i}}{\partial{W}_{i}} = \frac{\partial{L}_{i}}{\partial{W}} \odot {C}_{i}.
	\end{split}
	\end{equation}
	
	Parameters ${W}$ should fit all the individuals, and thus the gradients for all networks are accumulated to calculate the gradient of parameters ${W}$
	\vspace{-1em}
	\begin{equation}
	\label{eq_de_w}
	\begin{split}
	{dW} &= \frac{1}{P}\sum_{i=1}^{P}{dW}_{i} = \frac{1}{P}\sum_{i=1}^{P}\frac{\partial{L}_{i}}{\partial{W}} \odot {C}_{i}.
	\end{split}
	\end{equation}
	
	Any layer is only optimized by networks that use this layer during forwarding. By collecting the gradients of individuals in the population, the parameters ${W}$ are updated through the SGD algorithm.
	
	As we have maintained a large set of architectures with shared weights in the SuperNet, we borrow the idea of stochastic gradient descent and use mini-batch architectures for updating parameters. Accumulating the gradients for all networks would take much time for one-step gradient descent, and thus we use mini-batch architectures for updating shared weights. We use $B$ different architectures where $ B<P$, and the indices of architectures are $\{n_1, \dots, n_{B}\}$ to update parameters. The efficient parameter updating of Eqn~\ref{eq_de_w} is detailed as Eqn~\ref{eq_de_w_minibatch}
	\vspace{-1em}
	\begin{equation}
	\label{eq_de_w_minibatch}
	{dW} \approx \frac{1}{B}\sum_{j=1}^{B}\frac{\partial{L}_{n_j}}{\partial{W}_{n_j}}.
	\end{equation}
	
	Hence, the gradients over a mini-batch of architectures are taken as an unbiased approximation of the averaged gradients of all the $P$ different individuals. The time cost for each update could be largely reduced, and the appropriate mini-batch size leads to a balance between efficiency and accuracy.
	
	\begin{figure*}[t!]
		\centering
		%\begin{tabular}{cc}
		%  \centering
		%  \includegraphics[width=0.3\linewidth]{nsga3_aos.png} & \includegraphics[width=0.3\linewidth]{mars_nsga_aos.png} \\
		%  NSGA-III & pNSGA-III \\
		%\end{tabular}
		\subfigure[NSGA-III (1-5 generation)]{\label{fig_nsga3_aos}\includegraphics[width=0.24\linewidth]{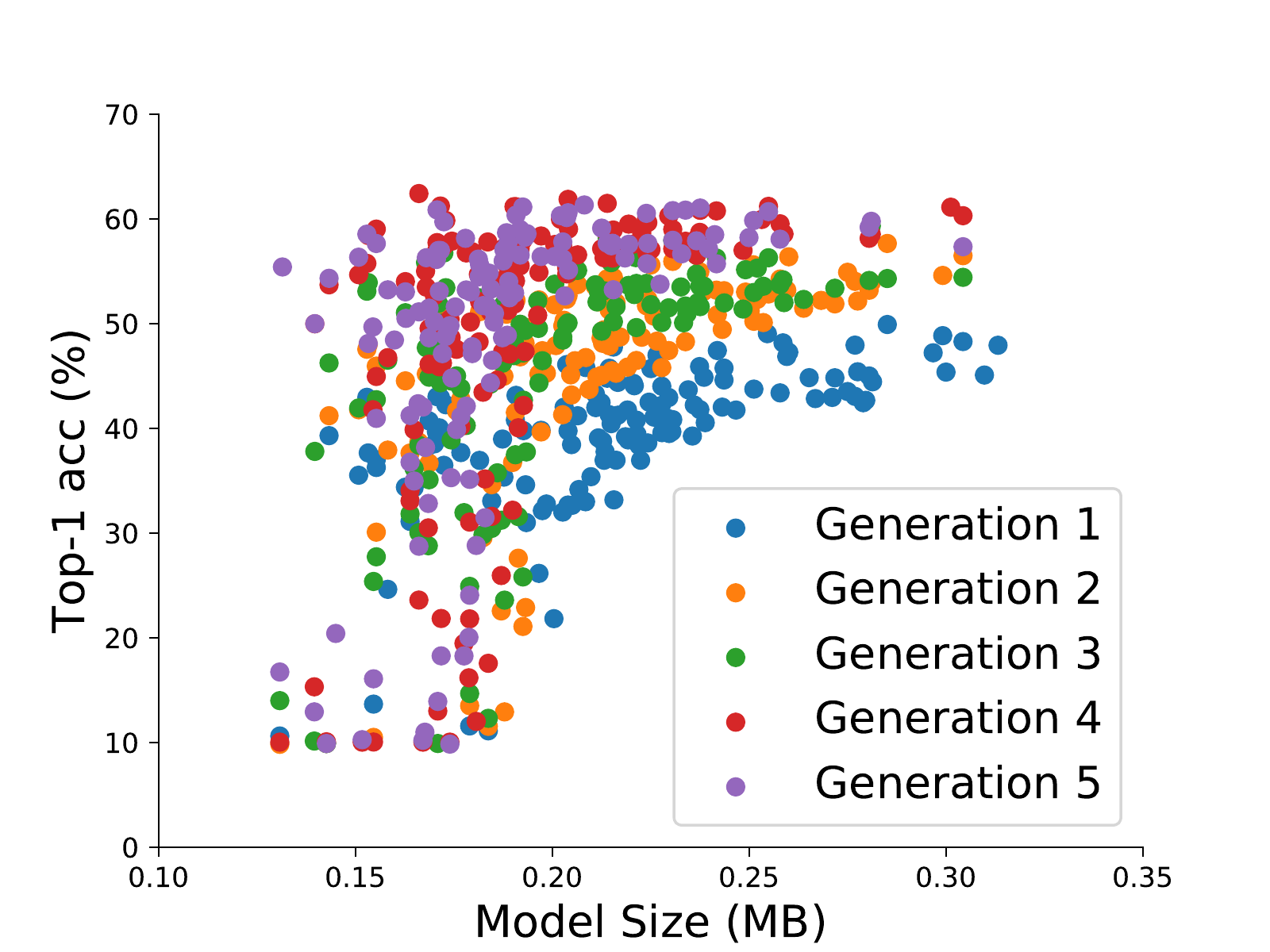}}
		\subfigure[NSGA-III (20 generation)]{\label{fig_nsga3_aos_sup}\includegraphics[width=0.24\linewidth]{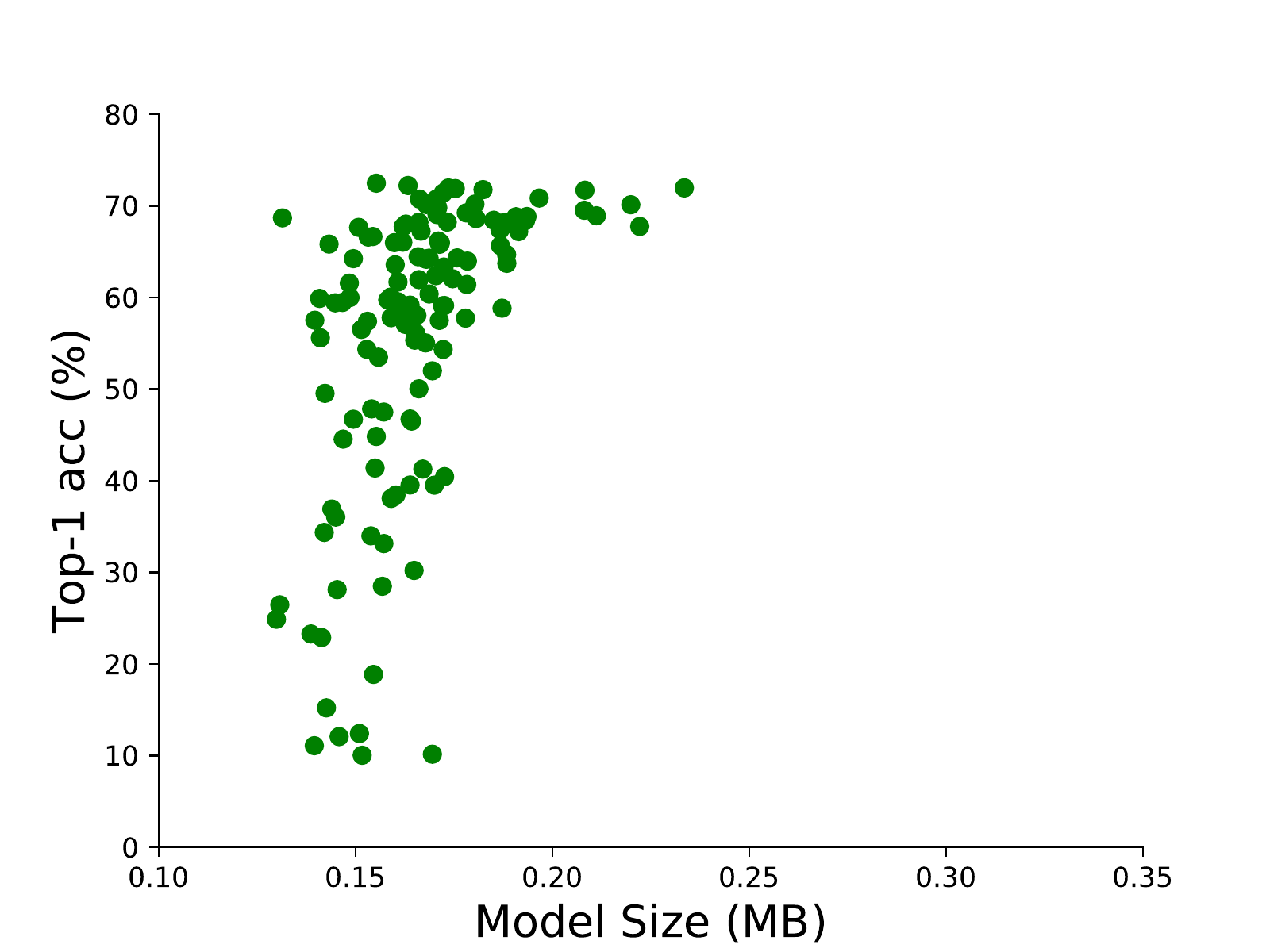}}
		\subfigure[pNSGA-III (1-5 generation)]{\label{fig_mars_nsga_aos}\includegraphics[width=0.24\linewidth]{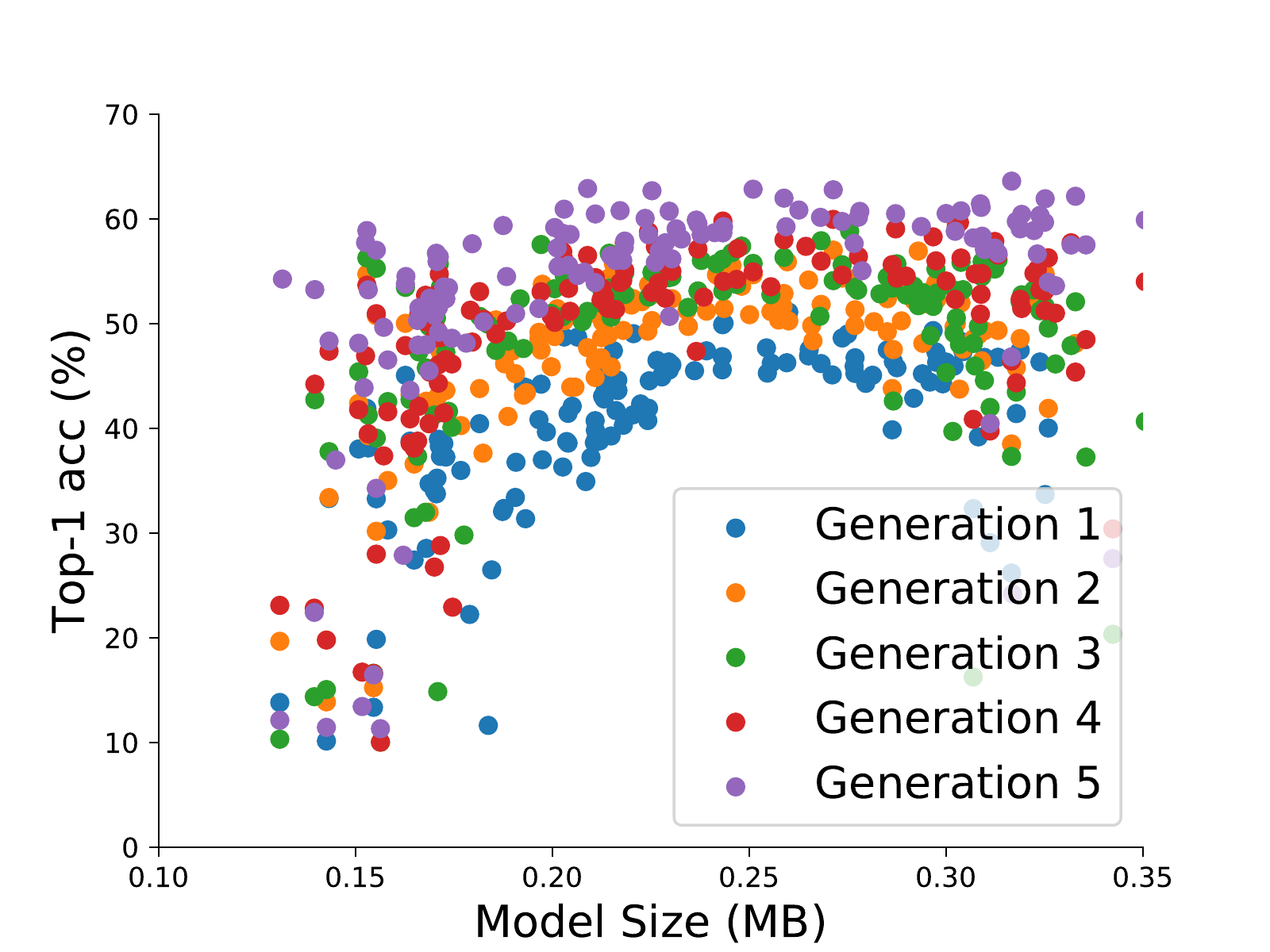}}
		\subfigure[pNSGA-III (20 generation)]{\label{fig_mars_nsga_aos_sup}\includegraphics[width=0.24\linewidth]{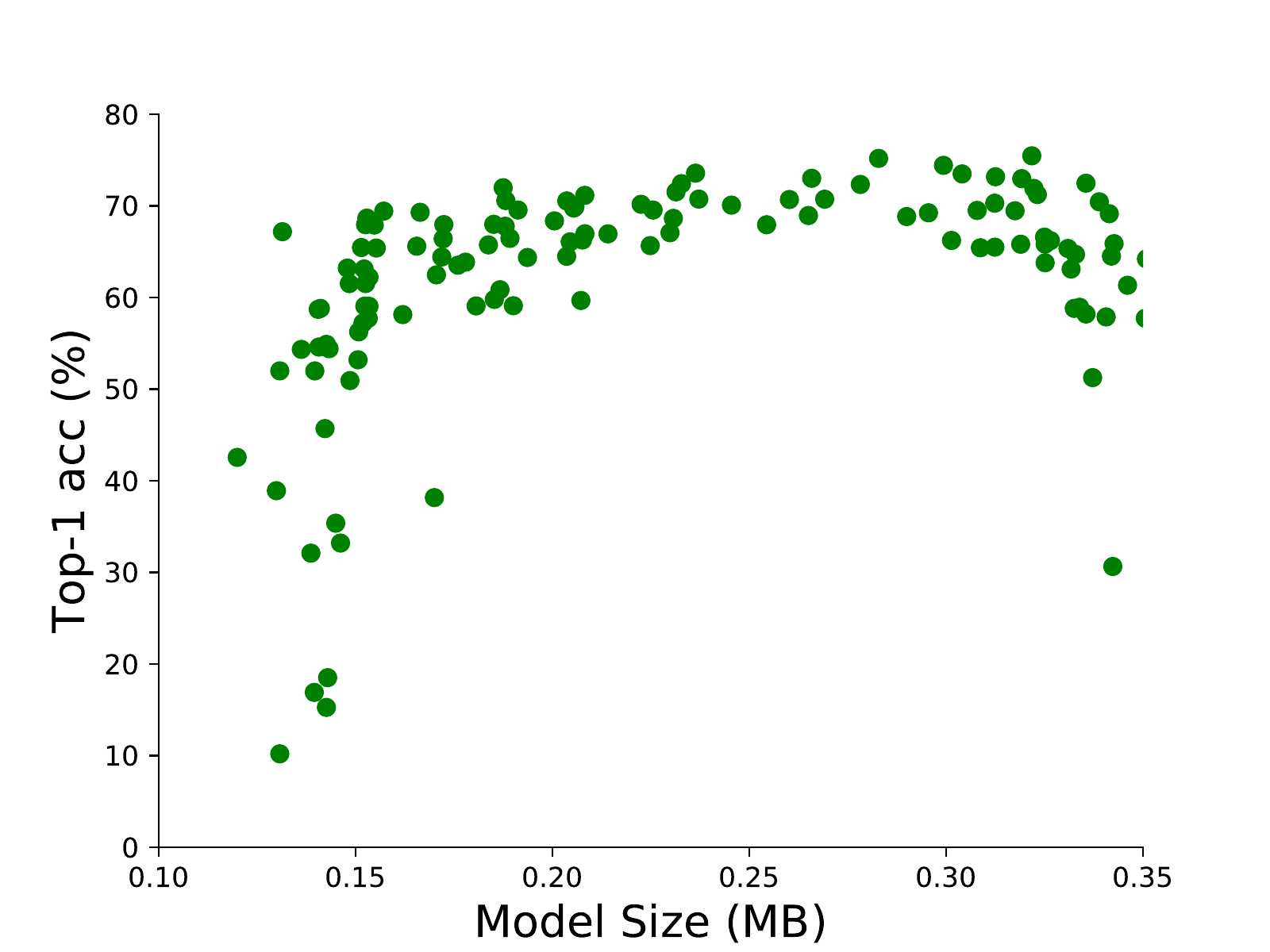}}
		\caption{Comparision between different evolution strategies. SuperNet is trained on the train set and evaluated on the validation set. Figure~\ref{fig_nsga3_aos} shows five evolution generations using NSGA-III. Evolution with NSGA-III suffers from small model trap, which lead the distribution bias to smaller models, and Figure~\ref{fig_nsga3_aos_sup} shows the distribution of evolution with NSGA-III after 20 generations, where the maintained architectures are all the small models. Figure~\ref{fig_mars_nsga_aos} shows the evolution generations using proposed pNSGA-III. Evolution with pNSGA-III provides protection for larger models, and Figure~\ref{fig_mars_nsga_aos_sup} shows the distribution of evolution with pNSGA-III after 20 generations, where the maintained architectures covers a large range over the model size dimension.}
		\label{fig_nsga}
		\vspace{-1em}
	\end{figure*}
	
	\subsection{Architecture Optimization}
	As for the architecture optimization procedure, we use the evolution algorithm together with the non-dominated sorting strategy. The non-dominated sorting strategy has been introduced in the NSGA-III~\cite{nsga3}. Denote $\{\mathcal{N}_{1}, \dots, \mathcal{N}_{P}\}$ as $P$ different networks and $\{\mathcal{F}_{1}, \dots, \mathcal{F}_{M}\}$ as $M$ different measurements we want to minimize. The measurements, for example, the number of parameters, float operations, latency, energy, and accuracy, could have some conflicts, which increase the difficulty in discovering an optimal solution that minimizes all these metrics.
	
	In practice, $\mathcal{N}_i$ dominates $\mathcal{N}_j$ if two conditions are satisfied: (1) For any of the measurements, the performance of $\mathcal{N}_i$ is not worse than that of $\mathcal{N}_j$. (2) The model $\mathcal{N}_i$ behaves better than $\mathcal{N}_j$ on at least one measurement. Formally, the definition of domination can be summarized as below.
	
	\theoremstyle{definition}
	\begin{definition}
		
		Considering two networks $\mathcal{N}_i$ and $\mathcal{N}_j$, and a series of measurements $\{\mathcal{F}_{1}, \dots, \mathcal{F}_{M}\}$ we want to minimize. If

		\begin{equation}
		\begin{aligned}
		\mathcal{F}_{k}(\mathcal{N}_{i}) & \leq \mathcal{F}_{k}(\mathcal{N}_{j}),\quad \forall k \in \{1,\dots,M\}\\
		\mathcal{F}_{k}(\mathcal{N}_{i}) & < \mathcal{F}_{k}(\mathcal{N}_{j}),\quad \exists k \in \{1,\dots,M\},
		\end{aligned}
		\end{equation}
	$\mathcal{N}_i$ is said to dominate $\mathcal{N}_j$, \ie, $\mathcal{N}_i \preceq \mathcal{N}_j$.
	\end{definition}
		
	According to the above definition, if $\mathcal{N}_i$ dominates $\mathcal{N}_j$, $\mathcal{N}_{j}$ can be replaced by $\mathcal{N}_{i}$ during the evolution procedure since $\mathcal{N}_{i}$ performs better in terms of at least one metric and not worse on other metrics. By exploiting this approach, we can select a series of excellent neural architectures from the population in the current generation. Then, these networks can be utilized for updating the corresponding parameters in the SuperNet.
	
\begin{figure*}
		\centering
		\subfigure[Acc curve (epochs=500, window=1)]{\label{fig_small_model_trap_left}\includegraphics[height=1.5in]{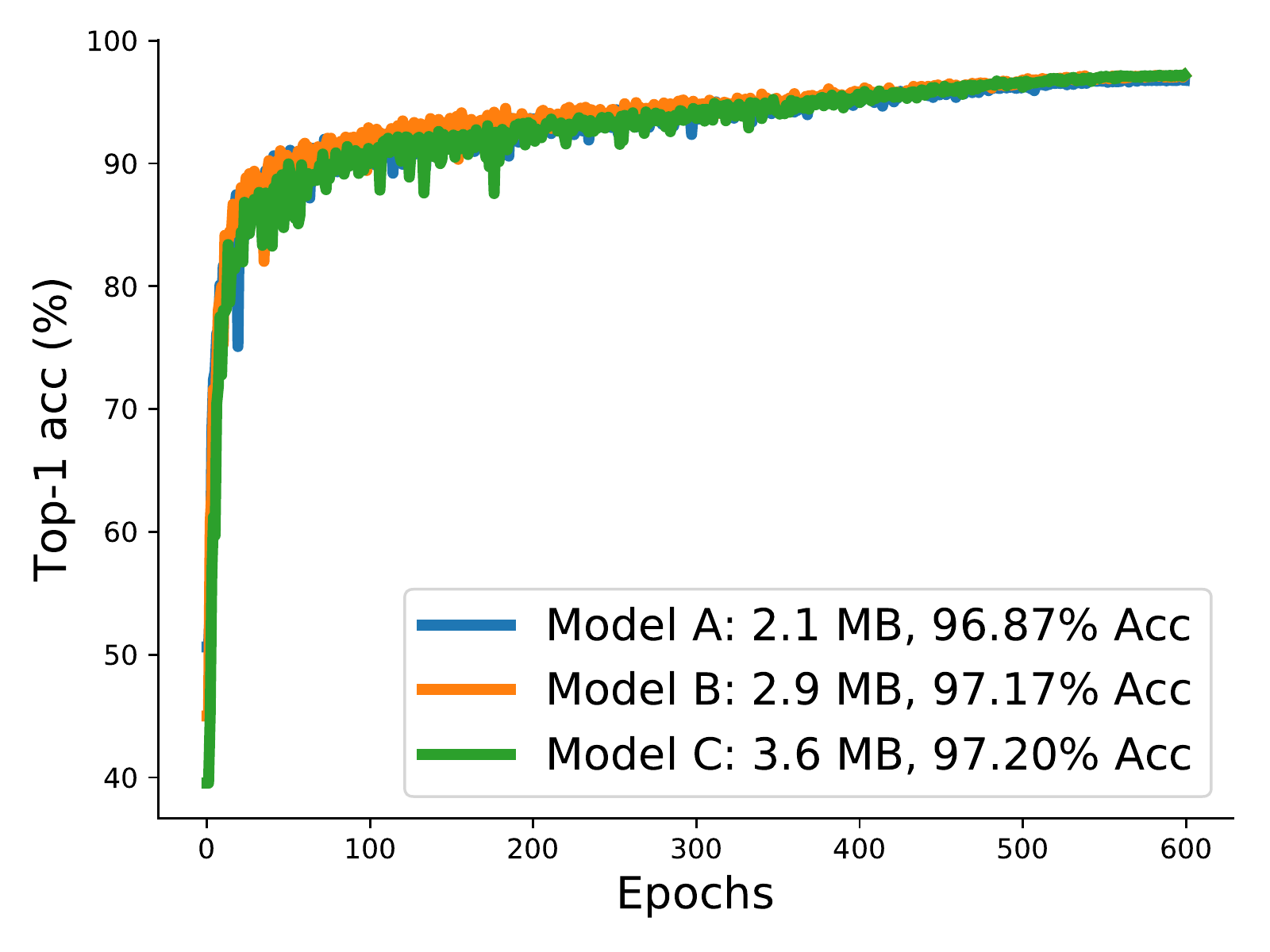}}
		\subfigure[Acc curve (epochs=50, window=1)]{\label{fig_small_model_trap_middle}\includegraphics[height=1.5in]{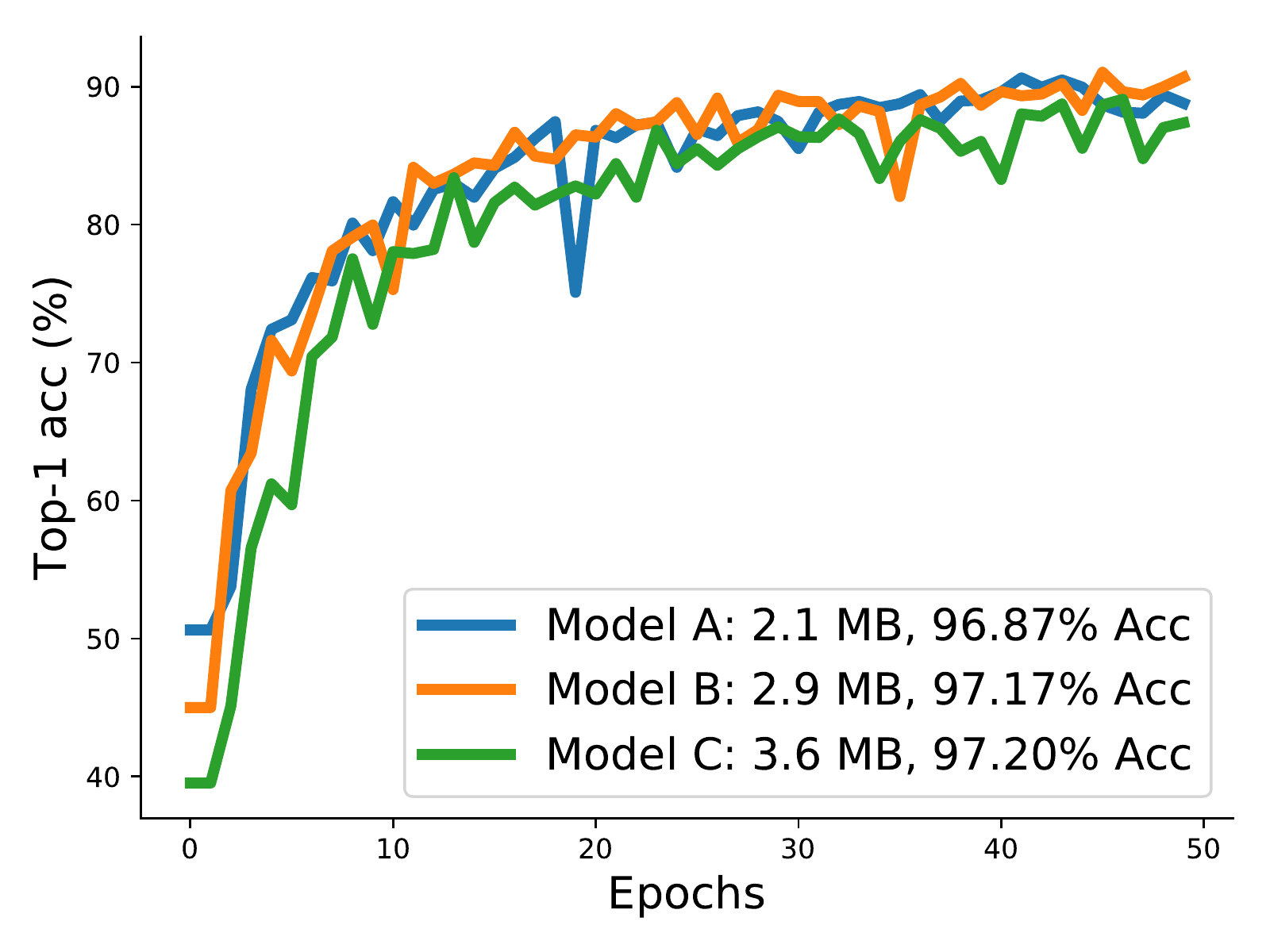}}
		\subfigure[Acc curve (epochs=50, window=5)]{\label{fig_small_model_trap_right}\includegraphics[height=1.5in]{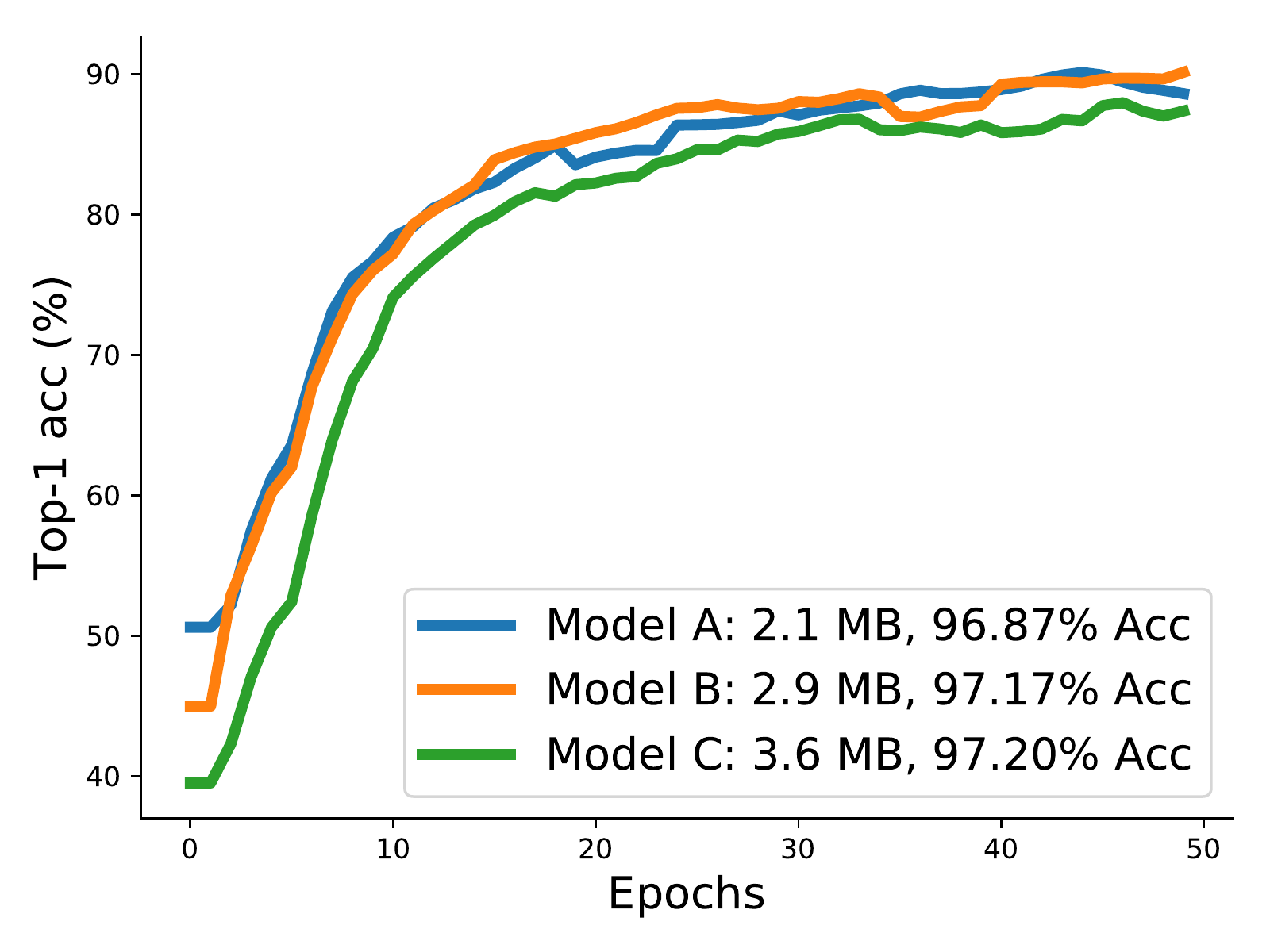}}
		\caption{The accuracy curves of three models with different model sizes. The left figure shows the accuracy curves of training for 600 epochs, the middle figure shows the accuracy curves for the first 50 epochs, and the right figure shows the curves smoothed by a window size of 5.}
		\label{fig_small_model_trap}
		\vspace{-1em}
	\end{figure*}

	Although the above non-dominated sorting strategy uses the NSGA-III method~\cite{nsga3} to select some better models for updating parameters, there exists \emph{a small model trap phenomenon} during the search procedure. Specifically, since the parameters in the SuperNet still need optimization, the accuracy of each individual architecture in the current generation may not always stand for its performance that can be eventually achieved, as discussed in NASBench-101~\cite{nasbench101}. Thus, some smaller models with fewer parameters but higher test accuracy tend to dominate those larger models of lower accuracy but have the potential for achieving higher accuracies, as shown in Figure~\ref{fig_small_model_trap}.
	
	Therefore, we propose to improve the conventional NSGA-III for protecting these larger models, namely pNSGA-III. More specifically, the pNSGA-III algorithm takes the increasing speed of the accuracy into consideration. We take the validation accuracy and the number of parameters as an example. For NSGA-III method, the non-dominated sorting algorithm considers two different objectives and selects individuals according to the sorted Pareto stages. For the proposed pNSGA-III, besides considering the number of parameters and accuracy, we also conduct a non-dominated sorting algorithm that considers the increasing speed of the accuracy and the number of parameters. Then the two different Pareto stages are merged.\footnote{The concerns about better middle sized models: The pNSGA-III process two times. For the first time, non-dominated sorting considers model size and accuracy (larger models with higher accuracy and smaller models with lower accuracy, the left side sorting). For the second time, non-dominated sorting considers the model size and increasing speed of accuracy (larger models with slower accuracy increasing speed and smaller models with faster accuracy increasing speed, the right side sorting). For the better middle-sized models, because they are larger and have higher accuracy than the small models, so they are ranked in the first few Pareto fronts for the first run. Also, the better middle-sized model with higher precision has a higher accuracy increasing speed than large models. Thus they are also ranked in the first few Pareto fronts for the second run. The selected architectures to construct the next generation is a hat-like form. So our proposed pNSGA-III will keep the accurate middle sized models in the next generation without omissions.} Assuming $P$ is the population size, after having two Pareto stages $R_{1 \dots n_1}, Q_{1 \dots n_2}$, we gradually merge two Pareto stages from the first Pareto front, and the union set $U_i$ after merging the $i$-th front is $U_i = (R_{1} \cup Q_{1}) \cup \dots \cup (R_{i} \cup Q_{i})$. We keep the first $P$ individuals from $U_{\max(n_1, n_2)}$. In this way, the large networks with slower performance increasing speed could be kept in the population.
	
	In Figure~\ref{fig_nsga}, the populations of using NSGA-III and pNSGA-III are visualized. If we use NSGA-III to update architectures, the small model trap problem is encountered. It is obvious that the pNSGA-III can protect large models during evolution and provide a wide range of models. More detailed discussions are introduced in the following section.
	
	\subsection{Continuous Evolution for CARS}
	\label{sec_training_CARS}
	
	In summary, there are two steps for searching optimal architectures by using the proposed CARS pipeline, 1) Architecture Optimization 2) Parameter Optimization. In addition, the parameter warm-up is also introduced to update the parameters at first.
	\paragraph{Parameter Warmup.} Since the shared weights of our SuperNet are randomly initialized, if the architectures in the population are also randomly initialized, the most frequently used operations for all the architectures would be trained more times compared with other operations. Thus, by following one-shot NAS methods~\cite{understandingoneshotnas,SPOSNAS,fairnas,fbnet}, we use a uniform sampling strategy to initialize the parameters in the SuperNet. In this way, the SuperNet trains each possible operation with the same possibility. For example, in DARTS~\cite{darts} pipeline, there are eight different operations for each node, including convolution, pooling, identity mapping and no connection. Each operation will be sampled with a probability of $\frac{1}{8}$.
	\paragraph{Architecture Optimization.} After initializing the parameters of the SuperNet, we first randomly sample $P$ different architectures, where $P$ is a hyper-parameter and denotes the number of maintained individuals in the population. During the architecture evolution step, we first generate $t \times P$ offsprings, where $t$ is the hyper-parameter controlling the expand ratio. We then use pNSGA-III to sort the architectures and select $P$ individuals from $ (t+1) \times P$ individuals. The selected $P$ architectures form the next generation.
	\paragraph{Parameter Optimization.} Given a set of architectures, we use the proposed mini-batch architectures update scheme for parameter optimization according to Eqn~\ref{eq_de_w_minibatch}.
	
	Algorithm~\ref{alg_train_CARS} summarizes the detailed procedure of the proposed continuous evolutionary algorithm for searching neural architectures.
	
	\subsection{Search Time Analysis}
	\label{sec_discussion}

	During the search stage of CARS, the train set is used for updating network parameters, and the validation set is used for updating architectures. Assuming the average training time on the train set for one architecture is $T_{tr}$, and the inference time on the validation set is $T_{val}$. The first warmup stage takes $E_{warm}$ epochs, and it needs $T_{warm} = E_{warm} \times T_{tr}$ in this stage to initialize parameters in the SuperNet $\mathcal{N}$. 
	
	Assuming the architectures evolve for $E_{evo}$ generations in total. And each generation contains parameter optimization and architecture optimization steps. The parameter optimization step trains the SuperNet for $E_{param}$ epochs on train set between generations, thus the time cost for parameter optimization in one evolution generation is $T_{param} = E_{param} \times T_{tr} \times B$, and the $B$ is the mini-batch size. For the architecture optimization step, all the individuals can be inferred in parallel, so the time cost in this step could be calculated as $T_{arch} = T_{val}$. Thus the total time cost for $E_{evo}$ evolution generations is $T_{evo} = E_{evo} \times ( T_{param} + T_{arch} )$. All the searching time cost in CARS is,
	\begin{equation}
	\begin{aligned}
	T_{total} = & T_{warm} + T_{evo}.\\
	= & E_{warm} \times T_{tr} + \\ & E_{evo} \times ( E_{param} \times T_{tr} \times B + T_{val} ) \\
	\end{aligned}
	\end{equation}
	
	\begin{algorithm}[t]
		\caption{Continuous Evolution for Efficient Neural Architecture Search}
		\begin{algorithmic}[1]
			\REQUIRE SuperNet $\mathcal{N}$, connections ${C} = \{{C}_1^{(0)}, \dots, {C}_P^{(0)}\}$, offspring expand ratio $t$, evolution number $E_{evo}$, multi-objectives $\{\mathcal{F}_{1}, \dots, \mathcal{F}_{M}\}$, parameter optimization epochs $E_{param}$, and criterion $\mathcal{H}$.
			\STATE Warm up the SuperNet $\mathcal{N}$ for $E_{warm}$ epochs.
			\FOR{ $e = 1, \dots, E_{evo}$}
			\FOR{ $i = 1, \dots, E_{param}$ }
			\FOR{Mini-batch data ${X}$, target ${Y}$ in loader}
			\STATE Random sample $B$ indices $n_1, \dots, n_B$.
			\STATE Select the corresponding $B$ connections ${C}_1, \dots, {C}_{n_B}$ according to the indices.
			\STATE Mask the SuperNet $\mathcal{N}$ to form the sampled networks $\mathcal{N}_1, \dots, \mathcal{N}_{n_B}$.
			\STATE Forward $B$ sampled networks.
			\STATE Calculate loss ${L}$ = $\frac{1}{B} \sum_{i=1}^{B}$ $\mathcal{H}$ ($\mathcal{N}_{n_i}({X})$, ${Y}$).
			\STATE Compute the gradients according to Eqn~\ref{eq_de_w_minibatch}.
			\STATE Update the network parameters $W$.
			\ENDFOR
			\ENDFOR
			\STATE Update $\{{C}_1^{(e)}, \dots, {C}_P^{(e)}\}$ using pNSGA-III.
			\ENDFOR
			\ENSURE Architectures ${C} = \{{C}_1^{(E_{evo})}, \dots, {C}_P^{(E_{evo})}\}$.
		\end{algorithmic}
		\label{alg_train_CARS}
	\end{algorithm}
	
		\begin{table*}
		\caption{Comparison with state-of-the-art image classifiers on CIFAR-10 dataset. The multi-objectives used for architecture optimization are performance and model size. We follow DARTS and use the cutout strategy for training.}
		\centering
		\small
		\begin{tabular}{l|c|c|c|c}
			\hline\hline
			{{Architecture}} & {Test Error (\%)} & {Params (M)} & {Search Cost (GPU days)} &  {Search Method} \\
			\hline
			DenseNet-BC~\cite{densenet} & 3.46 & 25.6 & - &  manual \\
			\hline
			PNAS~\cite{pnas} & 3.41 & 3.2 & 225 &  SMBO \\
			ENAS + cutout~\cite{enas} & 2.91 & 4.2 & 4 &  RL \\
			NASNet-A + cutout~\cite{nasnet} & 2.65 & 3.3 & 2000 &  RL \\
			AmoebaNet-A + cutout~\cite{amoebanet} & 3.12 & 3.1 & 3150 &  evolution \\
			Hierarchical evolution~\cite{hierarchical} & 3.75 & 15.7 & 300 &  evolution \\
			SNAS (mild) + cutout~\cite{snas} & 2.98 & 2.9 & 1.5 &  gradient\\
			SNAS (moderate) + cutout~\cite{snas} & 2.85 & 2.8 & 1.5 &  gradient \\
			SNAS (aggressive) + cutout~\cite{snas} & 3.10 & 2.3 & 1.5 &  gradient\\
			DARTS (first) + cutout~\cite{darts} & 3.00 & 3.3 & 1.5 &  gradient \\
			DARTS (second) + cutout~\cite{darts} & 2.76 & 3.3 & 4 &  gradient \\
			Random Search~\cite{darts} & 3.29 & 3.2 & 4 & random \\
			\hline
			RENA~\cite{rena} & 3.87 & 3.4 & - &  RL \\
			NSGANet~\cite{nsganet} & 3.85 & 3.3 & 8 &  evolution \\
			LEMONADE~\cite{lemonade} & 3.05 & 4.7 & 80 &  evolution \\
			\hline
			CARS-A & 3.00 & 2.4 & 0.4  &  evolution\\
			CARS-B & 2.87 & 2.7 & 0.4  &  evolution\\
			CARS-C & 2.84 & 2.8 & 0.4  &  evolution \\
			CARS-D & 2.95 & 2.9 & 0.4  &  evolution \\
			%CARS-DARTS-E & 2.93 & 3.0 & 0.4  & 7 & evolution \\
			CARS-E & 2.86 & 3.0 & 0.4  &  evolution \\
			CARS-F & 2.79 & 3.1 & 0.4  &  evolution  \\
			CARS-G & 2.74 & 3.2 & 0.4  &  evolution \\
			CARS-H & 2.66 & 3.3 & 0.4 &  evolution \\
			CARS-I & 2.62 & 3.6 & 0.4 &  evolution \\
			\hline\hline
		\end{tabular}
		\label{tab_CARS_darts}
	\end{table*}
	
	\section{Experiments}
	
	In this section, we first introduce the SuperNet, and experimental details in our experiments. Then, we examine the small model trap phenomenon, and compare  NSGA-III with our proposed pNSGA-III. We search on the CIFAR-10 dataset two times, which considers device-agnostic and device-aware objectives, respectively. All the searched architectures are evaluated on the CIFAR-10 and ILSVRC2012 dataset. These two datasets are the benchmarks for the recognition task.
		
	\subsection{Experimental Settings}
	
	%\paragraph{Datasets.} Our experiments are performed on the CIFAR-10 and ILSVRC2012~\cite{imagenet} datasets.  %We search architectures on the CIFAR-10 dataset, and the searched architectures are evaluated on the CIFAR-10 and ILSVRC2012 datasets.
		
	\paragraph{SuperNet Backbones.} To illustrate the effectiveness of our method, we evaluate our CARS on a popular used search space same as DARTS~\cite{darts}. DARTS is a differentiable NAS system and searches for reduction and normal cells. The normal cell is used for the layers that have the same spatial size of input feature and output feature. The reduction cell is used for layers with downsampling on input feature maps. After searching for these two kinds of cells, the network is constructed by stacking a set of searched cells. The search space contains eight different operations, including four types of convolution, two kinds of pooling, skip connect, and no connection. 
	
	\vspace{-1em}
	
	\paragraph{Evolution Details.} In the DARTS search space, each intermediate node in a cell is connected with two previous nodes. Crossover and mutation are conducted on the corresponding nodes. Both crossover ratio and mutation ratio are set to 0.25, and we randomly generate new architectures with a probability of 0.5. For the crossover operation, each node has a ratio of 0.5 to crossover its connections, and for mutation operation, each node has a ratio of 0.5 to be randomly reassigned.
	
	\subsection{Experiments on CIFAR-10}
	
	Our experiments on CIFAR-10 include the demonstration of the small model trap phenomenon, the comparison of NSGA-III and pNSGA-III, the device-agnostic and device-aware search. The evaluation are conducted on the CIFAR-10 dataset and the large ILSVRC2012 dataset. 
	
	\paragraph{Small Model Trap.} In Figure~\ref{fig_small_model_trap}, the accuracy curves of three models are shown. The number of parameters are 2.1M, 2.9M, and 3.6M, respectively. After training for 600 epochs, the accuracies on the CIFAR-10 dataset and the model sizes have a positive correlation, which are 96.87\%, 97.17\%, and 97.20\%. We observe the accuracy curves of the first 50 epochs and conclude two main reasons that result in the small model trap phenomenon. The two reasons are (1) small models are naturally converge faster, and (2) accuracy fluctuates during training. For the largest Model-C, its accuracy is consistently lower than Model-A and Model-B in the first 50 epochs. Therefore, if the NSGA-III algorithm is used, the Model-C will be eliminated, which is the first motivation for our proposed pNSGA-III. This is because larger models are more complex, thus harder to be optimized. For Model-B and Model-A, the accuracy curves are similar~(Figure~\ref{fig_small_model_trap_right}). However, due to the accuracy fluctuation during training~(Figure~\ref{fig_small_model_trap_middle}), if the accuracy of Model-A is higher than Model-B in one epoch, the Model-B would be eliminated for the non-dominated sorting strategy, which is the second reason we proposed. Both two reasons may lead to the elimination of large models in the process of architecture update. Thus, the proposed pNSGA-III is necessary to address the small model trap problem.
	
	\paragraph{NSGA-III vs. pNSGA-III.} We use CARS to search for architectures with different NSGA methods during the architecture optimization step. The multi-objectives are the number of parameters and model size. We visualize the distribution trend of the architectures maintained in the population. As Figure~\ref{fig_nsga} shows, updating architectures by using the NSGA-III would encounter the small model trap problem, and large models are eliminated during architecture optimization step. In contrast, updating architectures by using the pNSGA-III protect larger models.  The larger models have the potential to increase their accuracies during later epochs but converge slower than small models at the beginning. It is essential to maintain larger models in the population rather than dropping them during the architecture optimization stage if the search target is to find models with various computing resources.
	
	\paragraph{Search on CIFAR-10.} We split the CIFAR-10 train set into two parts, \ie, 25,000 images for updating network parameters and 25,000 for updating architectures. The split strategy is the same as DARTS~\cite{darts} and SNAS~\cite{snas}. We search for 500 epochs in total, and the parameter warmup stage lasts for the first 10\% epochs (50). After that, we initialize the population, which maintains $128$ different architectures and gradually evolve them using proposed pNSGA-III. We use pNSGA-III to update architectures after the network parameters are updated for ten epochs.
	
	\iffalse
	\paragraph{Search Time analysis.} For the experiment of considering the model size and performance, the training time on the train set $T_{tr}$ takes around 1 minute, and the inference time on val set $T_{val}$ is around 5 seconds. For the first initialization stage, it trains for 50 epochs, so the time cost in this stage $T_{warm}$ takes around an hour. The continuous evolution algorithm evolves the architectures for $E_{evo}$ generations. For the architecture optimization stage in one generation, the parallel evaluation time is $T_{arch} = T_{val}$. For the parameter optimization stage, we set $B$ to be 1 in our experiments since different batch size does not affect the overall growth trend for individuals in the population which is discussed in supplementary material, and we train the SuperNet $\mathcal{N}$ for 10 epochs in one evolution generation. So the parameter optimization time $T_{param}$ is about 10 minutes. Thus the time cost for total evolution generation $T_{evo}$ is around 9 hours, and the total searching time $T_{total}$ is around 0.4 GPU day. For the experiment of considering the latency and performance, the running latency for each model is evaluated during architecture optimization step. Thus the searching time is around 0.5 GPU day.
	\fi
	
	\paragraph{Evaluate on CIFAR-10.} After finishing the CARS search stage, there are $N=128$ architectures maintained in the population. We evaluate some architectures that have the similar model sizes with previous works~\cite{darts,snas} for comparison. We retrain the searched architectures on the CIFAR-10 dataset. All the training parameters are the same as DARTS~\cite{darts}.
	
	We compare the searched architectures with state-of-the-arts in Table~\ref{tab_CARS_darts}. All the searched architectures can be found in the supplementary material\footnote{https://github.com/huawei-noah/CARS}. Our searched architectures have the number of parameters that vary from 2.4M to 3.6M on CIFAR-10 dataset, and the performances of these architectures are on par with the state-of-the-arts. Meanwhile, if we evolve architectures by using NSGA-III method rather than pNSGA-III, we could only search for a set of architectures with approximately 2.4M parameters without larger models, and the models perform relatively poor.
	
	Compared to previous methods like DARTS and SNAS, our method is capable of searching for architectures over a broad range of the searching space. The CARS-G achieves comparable accuracy with DARTS (second-order), resulting in an approximate 2.75\% error rate with smaller model size. Using the same 3.3M parameters as DARTS (second-order), our CARS-H achieves lower test error, 2.66\% vs. 2.76\%. For the small models, our searched CARS-A/C/D also achieve comparable results with SNAS. Besides, our large model CARS-I achieves a lower error rate 2.62\% with slightly more parameters. The overall trend from CARS-A to CARS-J is that the error rate gradually decreases while increasing the model size. These models are all Pareto solutions. Compared to other multi-objective methods like RENA~\cite{rena}, NSGANet~\cite{nsganet} and LEMONADE~\cite{lemonade}, our searched architectures also show superior performance over these methods.
	
	\begin{figure}[t]
		\centering
		\begin{tabular}{cc}
			\includegraphics[width=0.42\linewidth]{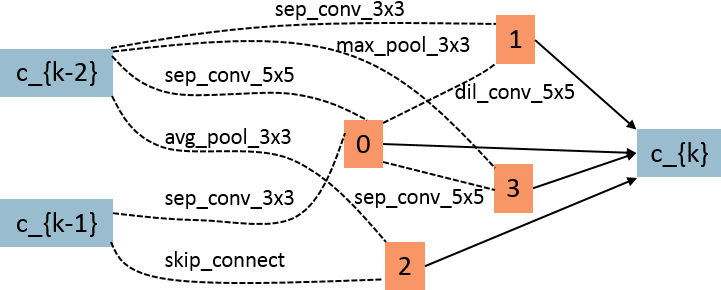}&
			\includegraphics[width=0.42\linewidth]{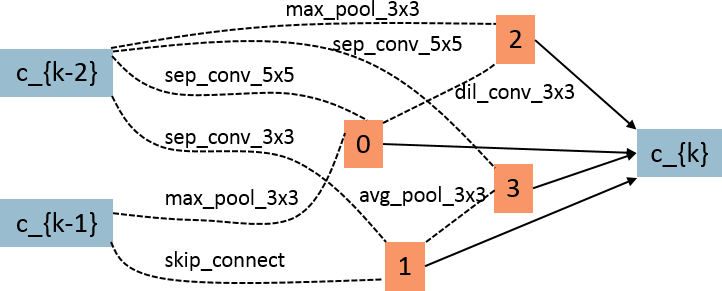}\\
			\includegraphics[width=0.42\linewidth]{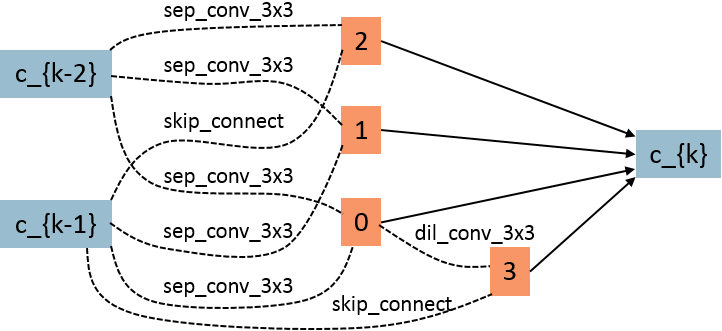}&
			\includegraphics[width=0.42\linewidth]{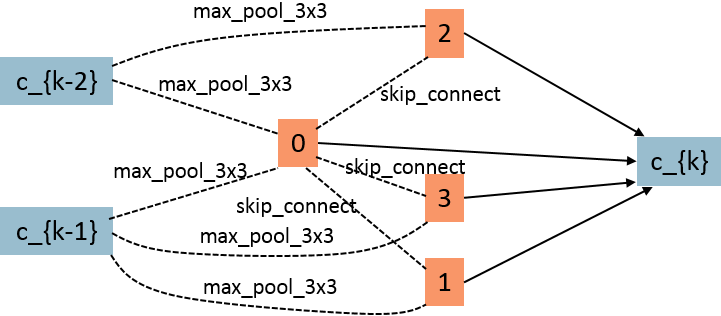}
		\end{tabular}
		\caption{CARS-H and DARTS. On the top are the normal and reduction blocks of CARS-H, and the bottom are the normal and reduction blocks in DARTS (second order). }
		\vspace{-1em}
		\label{fig_architecture_compare}
	\end{figure}
	
	\begin{table*}

		\caption{An overall comparison on ILSVRC2012 dataset. The CARS models are the architectures searched on the CIFAR-10 dataset.}
		\centering
		\small
		\begin{tabular}{l|c|c|c|c|c|c}
			\hline\hline
			\multirow{2}*{{Architecture}} & {Top-1} & {Top-5} & {Params} & {$+\times$} & {Search Cost} & {Search}\\
			~  & {Acc (\%)} & {Acc (\%)} & {(M)}  &     {(M)}    &    {(GPU days)} & {Method}  \\
			\hline
			ResNet50~\cite{resnet} & 75.3 & 92.2 & 25.6 & 4100 & - & manual\\
			MorphNet~\cite{morphnet} & 75.2 & - & 15.5 & 3880 & - & manual\\
			InceptionV1~\cite{googlenet} & 69.8 & 90.1 & 6.6 & 1448 & - & manual\\
			MobileNetV2 (1$\times$)~\cite{mobilenetv2} & 72.0 & 90.4 & 3.4 & 300 & - & manual\\
			%        MobileNetV2 (1.4$\times$)~\cite{mobilenetv2} & 74.7 & 92.5 & 6.9 & 585 & - & manual design\\
			ShuffleNetV2 (2$\times$)~\cite{shufflenetv2} & 74.9 & 90.1 & 7.4 & 591 & - & manual\\
			\hline
			PNAS~\cite{pnas} & 74.2 & 91.9 & 5.1 & 588 & 224 & SMBO\\
			AutoSlim~\cite{autoslim} & 75.4 & - & 8.3 & 532 & - & greedy \\
			SNAS (mild)~\cite{snas} & 72.7 & 90.8 & 4.3 & 522 & 1.5 & gradient \\
			DARTS~\cite{darts} & 73.3 & 91.3 & 4.7 & 574 & 4 & gradient\\
			PDARTS~\cite{pdarts} & 75.6 & 92.6 & 4.9 & 557 & 0.3 & gradient\\
			PARSEC~\cite{parsec} & 74.0 & 91.6 & 5.6 & 548 & 1 & gradient \\
			ProxylessNAS~(GPU)~\cite{proxylessnas} & 75.1 & 92.5 & 7.1 & 465 & 8.3 & gradient \\
			FBNet-C~\cite{fbnet} & 74.9 & - & 5.5 & 375 & 20 & gradient \\
			RCNet~\cite{rcnet} & 72.2 & 91.0 & 3.4 & 294 & 8 & gradient \\
			%MdeNAS~\cite{MdeNAS} & 74.5 & 92.1 & 6.1 & - & 0.16 & gradient\\
			NASNet-A~\cite{nasnet} & 74.0 & 91.6 & 5.3 & 564 & 2000 & RL\\
			%NASNet-B~\cite{nasnet} & 72.8 & 91.3 & 5.3 & 488 & 2000 & RL\\
			%NASNet-C~\cite{nasnet} & 72.5 & 91.0 & 4.9 & 558 & 2000 & RL\\
			MNasNet-A1~\cite{mnasnet} & 75.2 & 92.5 & 3.9 & 312 & - & RL \\
			%MNASNet-92~\cite{mnasnet} & 74.8 & 92.0 & 4.4 & 388 & 3800 & RL\\
			AmoebaNet-A~\cite{amoebanet} & 74.5 & 92.0 & 5.1 & 555 & 3150 & evolution\\
			%AmoebaNet-B~\cite{amoebanet} & 74.0 & 91.5 & 5.3 & 555 & 3150 & evolution\\
			%AmoebaNet-C~\cite{amoebanet} & 75.7 & 92.4 & 6.4 & 570 & 3150 & evolution\\
			\hline
			CARS-A & 72.8 & 90.8 & 3.7 & 430 & 0.4 & evolution \\
			CARS-B & 73.1 & 91.3 & 4.0 & 463 & 0.4 & evolution \\
			CARS-C & 73.3 & 91.4 & 4.2 & 480 & 0.4 & evolution \\
			CARS-D & 73.3 & 91.5 & 4.3 & 496 & 0.4 & evolution \\
			%CARS-DARTS-E & 73.5 & 91.5 & 4.4 & 502 & 0.4 & evolution + gradient \\
			CARS-E & 73.7 & 91.6 & 4.4 & 510 & 0.4 & evolution \\
			CARS-F & 74.1 & 91.8 & 4.5 & 530 & 0.4 & evolution \\
			CARS-G & 74.2 & 91.9 & 4.7 & 537 & 0.4 & evolution \\
			CARS-H & 74.7 & 92.2 & 4.8 & 559 & 0.4 & evolution \\
			CARS-I & 75.2 & 92.5 & 5.1 & 591 & 0.4 & evolution\\
			\hline\hline
		\end{tabular}
		\label{tab_params_acc}
	\end{table*}
	
	\paragraph{Comparison on Searched Cells.} In order to have an explicit understanding of the proposed method, we further visualize the normal and reduction cells searched by CARS and DARTS in Figure~\ref{fig_architecture_compare}. The CARS-H and DARTS (second-order) have a similar number of parameters (3.3M), but the CARS-H has higher accuracy compared to DARTS (second-order). It can be found in Figure~\ref{fig_architecture_compare}, there are more parameters in the CARS-H reduction block for preserving more useful information, and the size of the normal block of CARS-H is smaller than that of the DARTS (second-order) to avoid unnecessary computations. The CARS maintains a population that covers a large range of search space.

	\subsection{Evaluate on ILSVRC2012}
	
	We evaluate the transferability of the searched architectures by training them on the ILSVRC2012 dataset. We use 8 Nvidia Tesla V100 to train the models, and the batch size is 640. We train 250 epochs in total. The learning rate is 0.5 with a linear decay scheduler, and we warm up the learning rate for the first five epochs. Momentum is 0.9, and weight decay is 3e-5. Label smooth is also used with a smooth ratio of 0.1.
	
	The results in Table~\ref{tab_params_acc} show the transferability of our searched architectures. Our models cover an extensive range of parameters. The model sizes range from 3.7M to 5.1M, and the FLOPs range from 430 to 590 MFLOPs. For different deploy environments, we can easily select an architecture that satisfies the computing resources. This experiment considers the device-agnostic variables, model size and performance, thus the latencies of CARS A-I are not strictly positive related to the final performance. The latencies are 82.9, 83.3, 83.0, 90.0, 93.8, 92.2, 98.1, 97.2, 100.6 (ms) on HUAWEI P30 Pro. 
	
	The CARS-I surpasses PNAS by 1\% Top-1 accuracy with the same number of parameters and approximate FLOPs. The CARS-G shows superior results over DARTS by 0.9\% Top-1 accuracy with the same number of parameters. Also, CARS-D surpasses SNAS (mild) by 0.6\% Top-1 accuracy with the same number of parameters. For different models of NASNet and AmoebaNet, our method also has various models that achieve higher accuracy using the same number of parameters. By using the proposed pNSGA-III, the larger architectures like CARS-I could be protected during architecture optimization stages. Because of the efficient parameter sharing strategy, we could search a set of superior transferable architectures during the one-time search.
	
	For the experiment that considers device-aware variables, \ie, runtime latency and performance, we evaluate the searched architectures on the ILSVRC2012 dataset. The results are shown in Figure~\ref{fig_cars_darts_latency}. The searched architectures cover an actual runtime latency from 40ms to 90ms and surpass the counterparts.
	
	\section{Conclusion}
	
	The EA-based NAS methods are able to find models with high-performance, but the search time is extremely long because each candidate network is trained separately. In order to make this efficient, we propose a continuous evolution architecture search method, namely, CARS. During evolution, CARS maximally utilizes the learned knowledge in the latest evolution generation, such as architectures and parameters. A SuperNet is constructed with considerable cells and blocks. Individuals are generated through the benchmark operations in the evolutionary algorithm. The non-dominated sorting strategy (pNSGA-III) is utilized to select architectures for updating the SuperNet. Experiments on benchmark datasets show that the proposed CARS can efficiently provide several architectures on the Pareto front. The searched models are superior to the state-of-the-arts in terms of model size/latency and accuracy.	
	\newline
	
	\textbf{Acknowledgement}
	This work is supported by National Natural Science Foundation of China under Grant No. 61876007, 61872012,
Australian Research Council under Project DE180101438, and Beijing Academy of Artificial Intelligence~(BAAI).
	
	\clearpage
	\newpage
	\small
	\bibliographystyle{ieee_fullname}
	\bibliography{references}

\end{document}